\documentclass[10pt,twocolumn,letterpaper]{article}

\usepackage{wacv}              % To produce the CAMERA-READY version
\pdfoutput=1
\usepackage{graphicx}
\usepackage{amsmath}
\usepackage{amssymb}
\usepackage{booktabs}
\usepackage{amssymb}
\usepackage{pifont}
\usepackage{makecell}

\usepackage{enumitem}
\usepackage{arydshln}
\usepackage[accsupp]{axessibility} % Improves PDF readability for those with disabilities.

\usepackage[pagebackref,breaklinks,colorlinks]{hyperref}

\usepackage[capitalize]{cleveref}
\crefname{section}{Sec.}{Secs.}
\Crefname{section}{Section}{Sections}
\Crefname{table}{Table}{Tables}
\crefname{table}{Tab.}{Tabs.}

\def\wacvPaperID{2010} % *** Enter the WACV Paper ID here
\def\confName{WACV}
\def\confYear{2025}

\def\method{CATALOG}

\newcommand{\xmark}{\ding{55}} % X roja
\newcommand{\cmark}{\ding{51}} % Checkmark

\hypersetup{
  pdfauthor={Julian D. Santamaria, Claudia Isaza, Jhony H. Giraldo},
  pdftitle={CATALOG: A Camera Trap Language-guided Contrastive Learning Model},
  pdfkeywords={Camera Trap, Contrastive Learning, Computer Vision},
  pdfsubject={Proceedings of IEEE/CVF Winter Conference on Applications of Computer Vision, 2025}
}

\begin{document}

%\title{Language-guided CLIP Adaptation for Zero-shot Camera Trap Image Recognition}
%\title{Bridging the Gap between Foundation Models and Camera Trap Image Recognition}
\title{\method: A Camera Trap Language-guided Contrastive Learning Model}
\author{
Julian D. Santamaria$^1$, Claudia Isaza$^1$, Jhony H. Giraldo$^2$ \\
$^1$ SISTEMIC, Faculty of Engineering, Universidad de Antioquia-UdeA, Medellín, Colombia. \\
$^2$ LTCI, Télécom Paris, Institut Polytechnique de Paris, Palaiseau, France. \\
{\tt \small \{julian.santamaria, victoria.isaza\}@udea.edu.co, jhony.giraldo@telecom-paris.fr}
}

%\author{
%Julian D. Santamaria P\\
%SISTEMIC, Engineering Faculty \\
%Universidad de Antioquia-UdeA\\
%Medellín, Colombia\\
%{\tt\small julian.santamaria@udea.edu.co}
%\and
%Jhony H. Giraldo\\
%LTCI, Télécom Paris \\
%Institut Polytechnique de Paris\\
%Paris, France\\
%{\tt\small jhony.giraldo@telecom-paris.fr}
%\and
%Claudia Isaza\\
%SISTEMIC, Engineering Faculty \\
%Universidad de Antioquia-UdeA\\
%Medellín, Colombia\\
%{\tt\small victoria.isaza@udea.edu.co}
%}

\maketitle

%%%%%%%%% ABSTRACT
\begin{abstract}
   Foundation Models (FMs) have been successful in various computer vision tasks like image classification, object detection and image segmentation.
   However, these tasks remain challenging when these models are tested on datasets with different distributions from the training dataset, a problem known as domain shift.
   This is especially problematic for recognizing animal species in camera-trap images where we have variability in factors like lighting, camouflage and occlusions.
   In this paper, we propose the \textbf{Ca}mera \textbf{T}r\textbf{a}p \textbf{L}anguage-guided C\textbf{o}ntrastive Learnin\textbf{g} (\method) model to address these issues. 
   Our approach combines multiple FMs to extract visual and textual features from camera-trap data and uses a contrastive loss function to train the model.
   We evaluate \method~on two benchmark datasets and show that it outperforms previous state-of-the-art methods in camera-trap image recognition, especially when the training and testing data have different animal species or come from different geographical areas.
   Our approach demonstrates the potential of using FMs in combination with multi-modal fusion and contrastive learning for addressing domain shifts in camera-trap image recognition.
   The code of \method~is publicly available at \url{https://github.com/Julian075/CATALOG}.
   %The code of \method, provided in the supplementary, will be publicly released to facilitate further research and development in camera-trap image recognition.
\end{abstract}

%%%%%%%%% BODY TEXT
\section{Introduction}
\label{sec:intro}

In recent years, the field of deep learning has seen remarkable progress, driven in part by the emergence of a new class of models known as Foundation Models (FMs) \cite{xiao2024florence,radford2021learning,brown2020language,devlin2019bert,rombach2022high}.
These models are characterized by their large size, depth, and the vast amounts of data on which they have been trained, sometimes in the order of billions of data samples.
In computer vision, FMs have demonstrated exceptional performance in a wide range of tasks, including zero-shot image classification, object detection, and image segmentation \cite{fang2024simple,xu2023open,wu2023cora,radford2021learning}.
By leveraging the knowledge acquired during pre-training, FMs have enabled significant advances in the state-of-the-art of the classical computer vision tasks and opened up new possibilities for real-world applications.
One key advantage of some FMs, particularly those that incorporate text as an input modality, is their ability to handle open vocabulary tasks \cite{wu2024towards,du2022learning}.
These models are not restricted to a fixed set of categories but can recognize a wide range of objects based on descriptive text inputs \cite{pratt2023does,mall2022zero,zareian2021open}. 
This capability is particularly important in diverse and dynamic environments, such as wildlife monitoring, where the range of observable species can be extensive and unpredictable.

\begin{figure}
    \centering
    \includegraphics[width=\columnwidth]{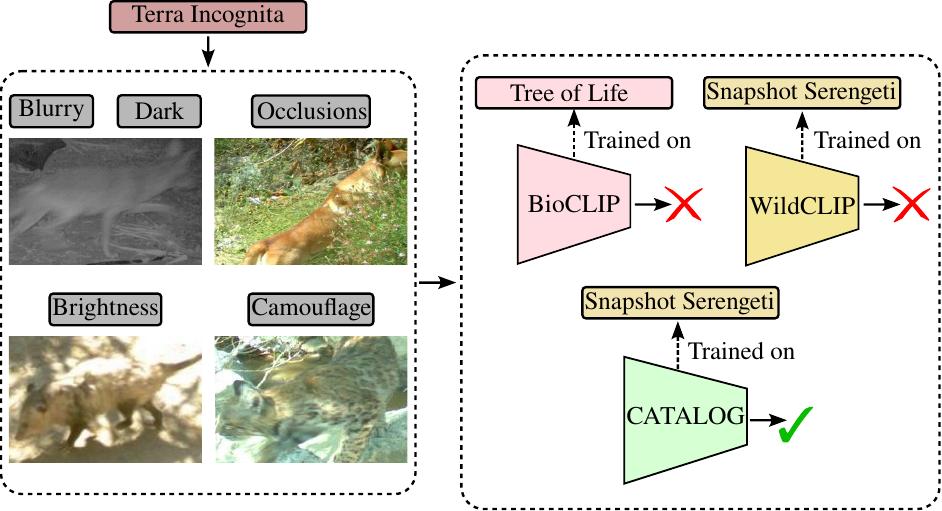}
    \caption{Comparison of CATALOG, BioCLIP \cite{stevens2024bioclip}, and WildCLIP \cite{gabeff2024wildclip} under challenging camera-trap conditions. CATALOG demonstrates superior performance.}
    %\caption{The performance of the foundation models for wildlife monitoring BioCLIP \cite{stevens2024bioclip} and WildCLIP \cite{gabeff2024wildclip} drops when evaluated on the Terra Incognita dataset \cite{beery2018recognition}.}
    \label{fig:teaser}
\end{figure}

Despite the success of FMs, adapting them to specific domains that differ significantly from the original training data remains a challenging task \cite{benigmim2024collaborating,oza2023unsupervised,yue2021prototypical}.
This difficulty is exacerbated when there is limited data, as it is often the case in specialized applications such as camera-trap image classification \cite{villa2017towards,beery2018recognition,schneider2020three,fabian2023multimodal}.
Camera traps are remote devices that are triggered by motion or heat to capture images or videos of wildlife in their natural habitat \cite{giraldo2019camera}.
While the use of camera traps has become increasingly popular in wildlife research and conservation efforts, collecting very large-scale datasets for this domain is still a significant challenge \cite{tuia2022perspectives,beery2018recognition,dryad_5pt92}.
%Because of the limited availability of data, it is necessary to explore alternative approaches that use existing models for camera-trap images.

% \begin{figure}
%     \centering
%     \includegraphics[width=1\columnwidth]{figures/winning_cases.pdf}
%     \caption{Comparison of CATALOG, BioCLIP \cite{stevens2024bioclip}, and WildCLIP \cite{gabeff2024wildclip} under challenging camera-trap conditions. CATALOG demonstrates superior robustness.}
%     \label{fig:winning_cases}
% \end{figure}

Due to these challenges, previous studies often adapt existing pre-trained FMs to the specific datasets instead of creating new models from scratch \cite{gabeff2024wildclip,stevens2024bioclip, fabian2023multimodal}.
For example, WildCLIP is an adaptation of the Contrastive Language-Image Pre-Training (CLIP) model \cite{radford2021learning}, which uses Vision Language Models (VLMs) tailored to camera-trap data \cite{gabeff2024wildclip}.
Similarly, BioCLIP adapts CLIP for the tree-of-life dataset to improve biological image analysis \cite{stevens2024bioclip}.
%is another adaptation of CLIP that is designed for the tree-of-life dataset to improve biological image analysis \cite{stevens2024bioclip}.
However, in the case of camera-trap images, we observe that: i) FMs often perform poorly on images that differ significantly from the training dataset, and ii) FMs do not generalize well in images that exhibit substantial variability in factors such as lighting, camouflage, and occlusions \cite{norman2023can,tuia2022perspectives,schneider2020three}.
These observations are exemplified in \cref{fig:teaser}.

% However, we have observed that these models often perform poorly on images that differ significantly from the training dataset, as illustrated in \cref{fig:teaser}.
% This limitation is particularly problematic for camera-trap images, which often exhibit substantial variability in factors such as lighting, pose, and occlusion, making it difficult for existing models to generalize effectively, as observed in \cref{fig:teaser} \cite{norman2023can,tuia2022perspectives,schneider2020three}.

In this paper, we introduce the \textbf{Ca}mera \textbf{T}r\textbf{a}p \textbf{L}anguage-guided C\textbf{o}ntrastive Learnin\textbf{g} (\method) model to recognize animal species in camera-trap images.
Our approach combines multiple FMs, including a Large Language Model (LLM) \cite{brown2020language}, CLIP \cite{radford2021learning}, LLaVA (Large Language-and-Vision Assistant) \cite{liu2024visual}, and BERT (Bidirectional Encoder Representations from Transformers) \cite{devlin2019bert}, to learn domain-invariant features from text and image modalities.
We introduce three key technical novelties: first, we combine text information from various sources using the centroid in the embedding space; second, we align the multi-modal features using a convex combination of the different sources; and third, we train our model using a contrastive loss to facilitate the learning of domain-invariant features for camera-trap images.
We train our model on the Snapshot Serengeti dataset \cite{dryad_5pt92} and evaluate it on the Terra Incognita dataset \cite{beery2018recognition}. The results demonstrate that \method~outperforms previous general-purpose and domain-specific FMs for camera-trap image recognition, especially when the domain of the training set differs from that of the testing set.
Our main contributions can be summarized as follows:
\begin{itemize}[leftmargin=*]
    \item We introduce a novel \method~model that integrates several FMs for camera-trap image recognition.
    \item When tested on datasets that differ from its training data, \method~outperforms previous FMs in recognizing animal species from camera-trap images.
    \item We conduct a series of ablation studies to confirm the effectiveness of each component in our model.
\end{itemize}

% The remainder of the paper is organized as follows:
% Sec. \ref{sec:related_works} presents an overview of the related work.
% Sec. \ref{sec:Method} provides a detailed explanation of the proposed \method~model.
% In Sec. \ref{sec:experiments_results}, we present the experimental framework, results, ablation studies, and discussion.
% Finally, Section \ref{sec:conclusions} offers the concluding remarks of the paper.

%\section{Related Work}

\section{Related Work}
\label{sec:related_works}

\noindent \textbf{Foundation models.}
In recent years, models trained with vast amounts of data, capable of learning high-level representations and performing complex tasks, have significantly advanced the fields of machine learning and artificial intelligence \cite{huang2024language,jiang2023mistral,touvron2023llama}. These models, powered by huge datasets, have achieved outstanding performance across various domains \cite{du2022survey}. Among these models, LLMs and VLMs are particularly remarkable for their exceptional ability to process images and text, while also generating coherent and relevant information \cite{zhang2023vision}.
%understand and generate text and images in a coherent and relevant manner. 
Notable examples of these models include GPT-3 \cite{brown2020language} and GPT-4 \cite{achiam2023gpt}. 
Due to their extensive pre-training, these LLMs have demonstrated the ability to generalize even in contexts for which they were not specifically trained \cite{miao2024new,fabian2023multimodal}.
Similarly, CLIP \cite{radford2021learning} has shown impressive results in image classification by learning joint representations of images and text. 
Another notable example of FM is the LLaVA model \cite{liu2024visual}, which integrates vision and language modalities to achieve state-of-the-art results in multi-modal tasks.

\vspace{0.1cm}

\noindent \textbf{Foundation models for biology.}
In biology, FMs have been adapted to address domain-specific challenges. For example, BioCLIP \cite{stevens2024bioclip} extends the principles of CLIP \cite{radford2021learning} to biological data, covering diverse categories such as plants, animals, and fungi.
BioCLIP also integrates rich structured biological knowledge. This model leverages the TREEOFLIFE-10M dataset \cite{stevens2024bioclip} and taxonomic names to achieve significant performance improvements in fine-grained classification tasks.
This enables the classification and analysis of complex biological images and text data.

\vspace{0.1cm}

\begin{figure*}
    \centering
    \includegraphics[width=\textwidth]{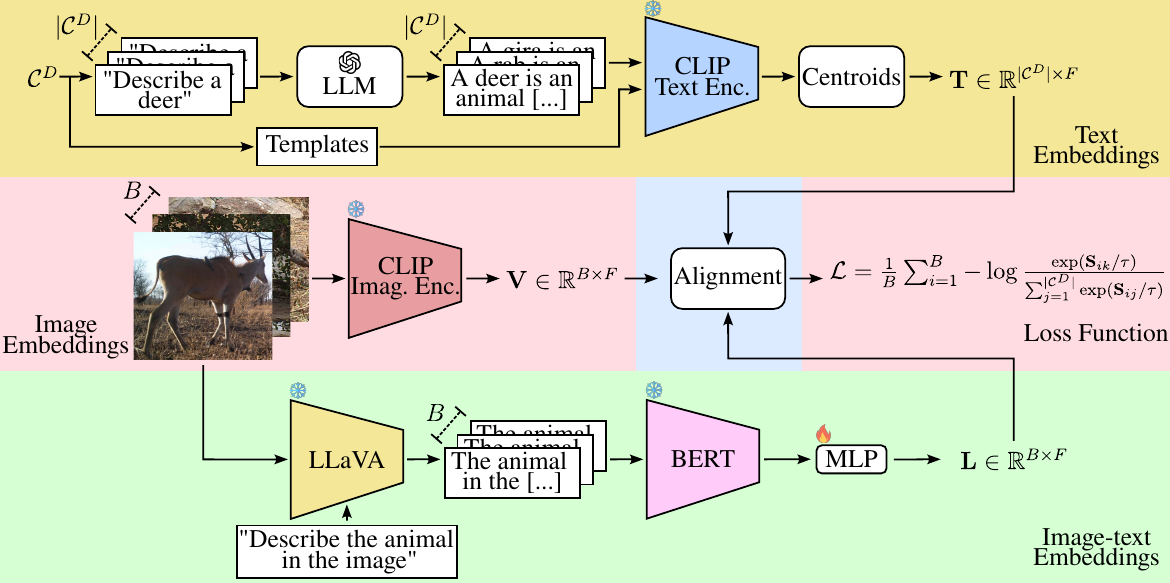}
    \caption{The pipeline of \method. Our model is divided into five parts: i) text embeddings, ii) image embeddings, iii) image-text embeddings, iv) feature alignment, and v) loss function. The textual embeddings are computed using a set of pre-defined templates and the LLM descriptions. The image embeddings are computed using CLIP. The image-text embeddings are calculated using LLaVA and BERT. Finally, we align the multi-modal features using an alignment mechanism and train the model with a contrastive loss function.}
    \label{fig:pipeline}
\end{figure*}

\noindent \textbf{Foundation models for camera traps.}
FMs are increasingly being applied to camera-trap data for wildlife monitoring and conservation. One such model is WildCLIP \cite{gabeff2024wildclip}, which uses the strengths of CLIP to accurately identify and classify different animal species in camera-trap images. Another approach is WildMatch \cite{fabian2023multimodal}, which introduces a zero-shot species classification framework. WildMatch adapts vision-language models to generate detailed visual descriptions of camera-trap images using expert terminology, which are then matched against an external knowledge base to identify species. These advances demonstrate the significant potential of FMs in improving wildlife monitoring and conservation efforts.

Previous methods for camera-trap image recognition have made progress in specific domains, but they often struggle when tested in diverse environmental contexts \cite{simoes2023deepwild} and under challenging conditions, as illustrated in \cref{fig:teaser}.
Our proposed model, \method, addresses this limitation by leveraging feature representation from FMs that are robust against domain shifts.
This integration improves species recognition and contextual understanding, making the model less sensitive to variations in environmental conditions and new classes.
% As shown in \cref{fig:teaser}, the examples illustrate how our approach effectively handles scenarios where previous methods struggle, showing enhanced adaptability and more accurate feature extraction.
% This leads to more comprehensive representations for camera-trap images and solves issues where other methods often fail.

\section{\method}
\label{sec:Method}

\noindent \textbf{Problem definition.}
In this paper, we assume access to an annotated training dataset, denoted as $\mathcal{D}$, which consists of $N_d$ image-label pairs, $\mathcal{D}=\{(\mathbf{x}_{i}^{D},\mathbf{y}_{i}^{D})\}_{i=1}^{N_d}$, with a set of classes $\mathcal{C}^{D}$.
For testing purposes, we have another dataset, $\mathcal{S}$, containing $N_s$ image-label pairs, $\mathcal{S} = \{ (\mathbf{x}_{i}^{S},\mathbf{y}_{i}^{S}) \}_{i=1}^{N_s}$, with a set of classes $\mathcal{C}^{S}$.
The sets of classes in both datasets may or may not overlap, meaning that $\mathcal{C}^{D} \cap \mathcal{C}^{S}$ may or may not be empty. 
Both datasets are derived from the natural world, but their images may not necessarily come from the same distribution, resulting in a domain shift.
Our goal is to train a deep learning model using only the training dataset $\mathcal{D}$ and deploy it on the testing dataset $\mathcal{S}$.

\subsection{Overview of the Approach}

\cref{fig:pipeline} presents the pipeline of our proposed framework, \method.
Our approach consists of three main components: i) text, ii) image, and iii) image-text embeddings.
For the text component, we input our dictionary of classes, $\mathcal{C}^{D}$, and use an LLM with predefined templates to generate descriptions for each category.
Then, we utilize CLIP's text encoder to obtain embeddings for each textual description.
To ensure a unique embedding for each class in $\mathcal{C}^{D}$, we apply a technique to combine the embeddings (Sec.~\ref{sec:Text Embeddings}), resulting in a single embedding of dimension $F$.
For the image component, we use CLIP's image encoder to extract embeddings from a mini-batch of $B$ images (Sec.~\ref{sec:Image Embeddings}).
Meanwhile, for the image-text component, we employ the VLM LLaVA coupled with BERT and a Multi-Layer Perceptron (MLP) to compute image-text embeddings from the mini-batch of images (Sec.~\ref{sec:Image-text Embeddings}).
To ensure the embeddings from all three components are aligned, we use an alignment mechanism (Sec.~\ref{sec:embeddings_aligmnet}). 
Finally, we utilize the output of the alignment mechanism to compute a contrastive loss, which is used to train our model (Sec.~\ref{sec:contrative}).
We let all FMs frozen (\raisebox{-.3ex}{\includegraphics[height=0.3cm]{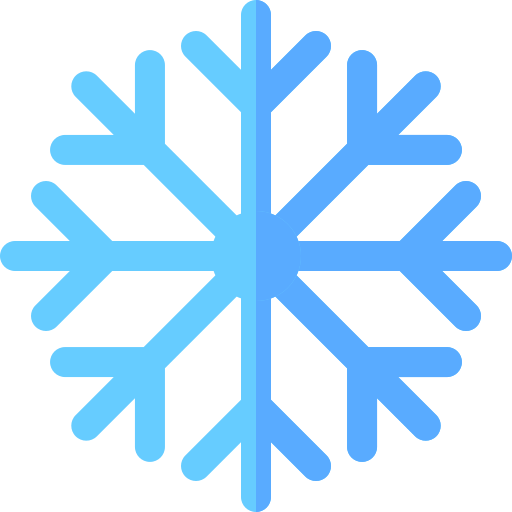}}) and only train (\raisebox{-.3ex}{\includegraphics[height=0.3cm]{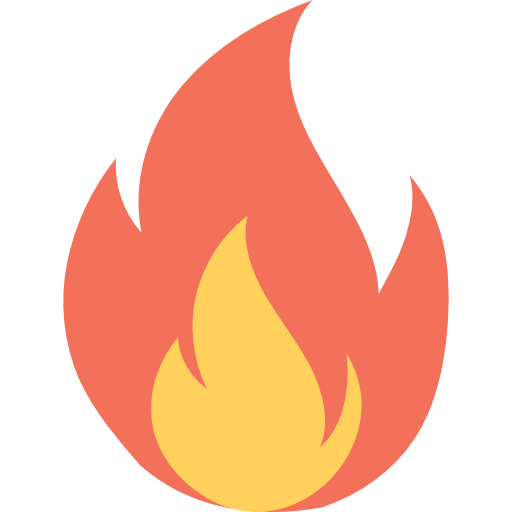}}) the MLP.

\subsection{Text Embeddings}
\label{sec:Text Embeddings}

To generate textual descriptions for each category in our dataset $\mathcal{C}^D$, we utilize an LLM that can provide detailed information about the animals without requiring expert inputs.
We also create multiple descriptions using predefined templates tailored to our specific task of camera-trap image recognition such as ``a photo captured by a camera trap of a \{ \}".
These templates add context to the descriptions by specifying that the images were captured by camera traps.
Examples of the precise prompts and templates used can be found in the supplementary material.
We process two types of textual descriptions using CLIP's text encoder: one generated by the LLM and $M-1$ manually crafted templates.
As a result, for each class in $\mathcal{C}^D$, we obtain $M$ embeddings, each with a dimension of $F$.
This allows us to represent each class using a set of textual embeddings that capture the semantic meaning of the category.

To obtain the final embedding for each class $c \in \mathcal{C}^D$, we compute the centroid of the $M$ embeddings generated for that class.
Specifically, let $\mathbf{P}^{(c)} \in \mathbb{R}^{M \times F}$ be the set of $M$ embedding for the class $c \in \mathcal{C}^D$.
The final embeddings $\mathbf{t}_c \in \mathbb{R}^F$ of $c$ is calculated as the average of these $M$ embeddings, as follows:
\begin{equation}
    \mathbf{t}_c = \frac{1}{M} \sum_{i=1}^M \mathbf{P}^{(c)}_{i:},
\end{equation}
where $\mathbf{P}^{(c)}_{i:}$ represents the $i$th row of $\mathbf{P}^{(c)}$.
The output of the text embedding part of \method~is a matrix $\mathbf{T} = [\mathbf{t}_1, \mathbf{t}_2, \dots,\mathbf{t}_{\vert \mathcal{C}^D \vert}]^\top \in \mathbb{R}^{\vert \mathcal{C}^D \vert \times F}$, which contains the final embeddings for all classes in $\mathcal{C}^D$.

% We compute the final embedding for each class $c \in \mathcal{C}^D$ as the centroid of the $M$ embeddings.
% In other words, let $\mathbf{P}^{(c)} \in \mathbb{R}^{M \times F}$ be the set of $M$ embedding for the class $c \in \mathcal{C}^D$.
% The final embedding $\mathbf{t}_c \in \mathbb{R}^F$ of $c$ is given by:
% \begin{equation}
%     \mathbf{t}_c = \frac{1}{M} \sum_{i=1}^M \mathbf{P}^{(c)}_{i:},
% \end{equation}
% where $\mathbf{P}^{(c)}_{i:}$ is the $i$th row of $\mathbf{P}^{(c)}$.
% The output of the text embedding part of \method~is a matrix $\mathbf{T} = [\mathbf{t}_1, \mathbf{t}_2, \dots,\mathbf{t}_{\vert \mathcal{C}^D \vert}]^\top \in \mathbb{R}^{\vert \mathcal{C}^D \vert \times F}$.

\subsection{Image Embedding}
\label{sec:Image Embeddings}

\noindent \textbf{Pre-processing.}
We utilize the MegaDetector model \cite{beery2019efficient} to process our camera trap datasets.
The purpose of using this model is to extract crops from the camera-trap images that contain relevant information. 

\vspace{0.1cm}

\noindent \textbf{CLIP embeddings.}
We use CLIP's image encoder \cite{radford2021learning} to extract embeddings from the cropped images. 
We process the images in mini-batches of size $B$.
%For each image in the mini-batch, we extract an embedding of dimension $F$ using the CLIP encoder. 
For each image, we extract an embedding of dimension $F$ using the CLIP encoder.
The output of this stage is a matrix $\mathbf{V} = [\mathbf{v}_1, \mathbf{v}_2, \dots, \mathbf{v}_B]^\top \in \mathbb{R}^{B \times F}$, where $\mathbf{v}_i$ corresponds to the visual embedding of the $i$th image in the mini-batch. 
This matrix is then used in the subsequent stages of our framework to align and contrast the text and image embeddings.

\subsection{Image-text Embeddings}
\label{sec:Image-text Embeddings}

In the image-text branch of \method, we use the mini-batch of cropped images as input. 
We employ the LLaVA model \cite{liu2024visual} to generate textual descriptions of the animals present in the cropped images, using a prompt similar to the one described in \cite{fabian2023multimodal} (provided in the supplementary material).
These textual descriptions are processed using the BERT model, which we selected because it provides 512 possible tokens \cite{yang2019emotionx}.
%CLIP's text input is limited to a maximum of 77 tokens \cite{ding2024clip,zhang2024long}.
%For our purposes, we need longer descriptions to accurately describe the input images.
We also utilize Long CLIP \cite{zhang2024long}, but the results are not satisfactory (see Sec. \ref{sec:ablations}).
%which we select because CLIP's text input is limited to a maximum of 77 tokens \cite{ding2024clip,zhang2024long}.
%For our purposes, we need longer descriptions to accurately describe the input images.
BERT generates text embeddings of dimension $F'$.
Since $F'$ is not equal to the dimension $F$ of the CLIP embeddings, we feed these BERT embeddings into an MLP to match the dimensions.

The MLP serves to project each BERT embedding into a $F$-dimensional space to match the CLIP embedding dimension. 
However, it does not perform the alignment between the two different embeddings.
The actual alignment between BERT and CLIP embeddings is achieved through the alignment mechanism and loss function, which we describe in detail in Sec.~\ref{sec:embeddings_aligmnet} and Sec.~\ref{sec:contrative}.

The output of the image-language branch of \method~is a matrix $\mathbf{L} = [\mathbf{l}_1, \mathbf{l}_2, \dots, \mathbf{l}_B]^\top \in \mathbb{R}^{B \times F}$, where $\mathbf{l}_i$ is the transformed BERT embedding of the $i$th image in the mini-batch.

\subsection{Alignment Mechanism}
\label{sec:embeddings_aligmnet}

The alignment method takes the text ($\mathbf{T}$), image ($\mathbf{V}$), and image-text ($\mathbf{L}$) embeddings from each branch as input.
The feature alignment process consists of two parts: i) similarity computation and ii) fusion mechanism.
For the similarity computation, we calculate the cosine similarities between text and image embeddings, as well as text and image-text embeddings. 
Specifically, let $\mathbf{W} \in \mathbb{R}^{B \times \vert \mathcal{C}^D \vert}$ be the matrix of cosine similarities between the text and image embeddings, computed as follows:
\begin{equation}
    \mathbf{W}_{ij} = \frac{\langle \mathbf{v}_i, \mathbf{t}_j \rangle}{\Vert \mathbf{v}_i\Vert \Vert \mathbf{t}_j \Vert}~\forall~1\leq i \leq B, 1\leq j \leq \vert \mathcal{C}^D \vert,
\end{equation}
where $\mathbf{W}_{ij}$ represents the $(i,j)$ item of the matrix, $\langle \cdot, \cdot \rangle$ denotes inner product, and $\Vert \cdot \Vert$ is the $\ell_2$ norm of a vector.
Similarly, we compute the cosine similarities between the text and image-text embeddings as follows:
\begin{equation}
    \mathbf{Q}_{ij} = \frac{\langle \mathbf{l}_i, \mathbf{t}_j \rangle}{\Vert \mathbf{l}_i\Vert \Vert \mathbf{t}_j \Vert}~\forall~1\leq i \leq B, 1\leq j \leq \vert \mathcal{C}^D \vert,
\end{equation}
where $\mathbf{Q} \in \mathbb{R}^{B \times \vert \mathcal{C}^D \vert}$ is the matrix of cosine similarities between the text and image-text embeddings.

The fusion mechanism is implemented as a weighted average between the matrices $\mathbf{W}$ and $\mathbf{Q}$, where the weights are determined by the hyperparameter $\alpha \in [0,1]$ .
Specifically, the output of the fusion method is a matrix $\mathbf{S} \in \mathbb{R}^{B \times \vert \mathcal{C}^D \vert}$, defined as follows:
\begin{equation}
    \mathbf{S} = \alpha \mathbf{W} + (1- \alpha) \mathbf{Q}.
    \label{eqn:fusion}
\end{equation}
Since $\alpha \in [0,1]$, the resulting matrix $\mathbf{S}$ is a convex combination of $\mathbf{W}$ and $\mathbf{Q}$.
This means that each element $\mathbf{S}_{ij}$ of the matrix is also between $0$ and $1$.

\subsection{Contrastive Loss}
\label{sec:contrative}

We train our model using a contrastive loss function, $\mathcal{L}$, which takes the matrix $\mathbf{S}$ as input.
The loss function is calculated for each mini-batch as follows:
\begin{equation}
    \mathcal{L}(\mathbf{S}) = \frac{1}{B} \sum_{i=1}^B -\log \frac{\exp(\mathbf{S}_{ik}/\tau)}{\sum_{j=1}^{\vert \mathcal{C}^D \vert} \exp(\mathbf{S}_{ij}/\tau)},
    \label{eqn:loss}
\end{equation}
where $\tau$ is a temperature hyperparameter and $k$ is the index of the class in $\mathcal{C}^D$ of the $i$th image in the mini-batch.
The intuition behind this loss function is to bring the image-text embeddings close to the text and image embeddings in the feature space when they correspond to the same species category in the dataset.
Conversely, we aim to push apart the multi-modal embeddings from other species in the mini-batch.
This encourages the model to learn a shared embedding space where the embeddings from different modalities are similar for the same species.

\section{Experiments and Results}
\label{sec:experiments_results}

This section presents the datasets used in the current work, the evaluation protocol, the implementation details, the results, and the discussion of \method. 
We compare our algorithm against CLIP \cite{radford2021learning}, BioCLIP \cite{stevens2024bioclip}, and WildCLIP \cite{gabeff2024wildclip}.
We also perform a set of ablation studies to analyze each component of \method, including the set of prompts and textual information we use to describe the different categories, different architectural choices like the VLM, CLIP's image encoder, and the loss function.
Finally, we study the sensibility of \method~regarding the hyperparameter $\alpha$ in the alignment mechanism.

\vspace{0.1cm}

\begin{table*}[t]
\centering
\def\arraystretch{0.9}
\resizebox{\textwidth}{!}{
\begin{tabular}{lccccc}
\toprule
\textbf{Model} & \textbf{Backbone} & \textbf{Training} & \textbf{Test} & \textbf{Cis-Test Acc (\%)} & \textbf{Trans-Test Acc (\%)}  \\ \midrule
CLIP \cite{radford2021learning}  & ViT-B/32 & OpenAI data & Terra Incognita & $32.18$ & $26.62$  \\ 
CLIP \cite{radford2021learning} & ViT-B/16 & OpenAI data & Terra Incognita & $39.14$ & $34.67$  \\ 
BioCLIP \cite{stevens2024bioclip} & ViT-B/16 & TREEOFLIFE-10M & Terra Incognita & $21.12$ & $14.53$ \\
WildCLIP \cite{gabeff2024wildclip} & ViT-B/16  & Snapshot Serengeti & Terra Incognita & $40.38$ & $38.90$  \\ 
WildCLIP-LwF \cite{gabeff2024wildclip} & ViT-B/16 & Snapshot Serengeti & Terra Incognita & $41.60$ & $36.20$  \\ \hdashline
\textbf{\method~(ours)} & ViT-B/16 & Snapshot Serengeti & Terra Incognita & $\textbf{48.59}$ & $\textbf{41.92}$ \\
\bottomrule
\end{tabular}
}
%}
\caption{Zero-shot performance results of \method~and other foundation models in the Terra Incognita dataset (out-of-domain evaluation). All methods are trained in data that differ from the test dataset. The best method is highlighted in \textbf{bold}.}
\label{tab:zero-shot-results}
\end{table*}

\begin{figure}[t]
    \centering
    \includegraphics[width=.9\columnwidth]{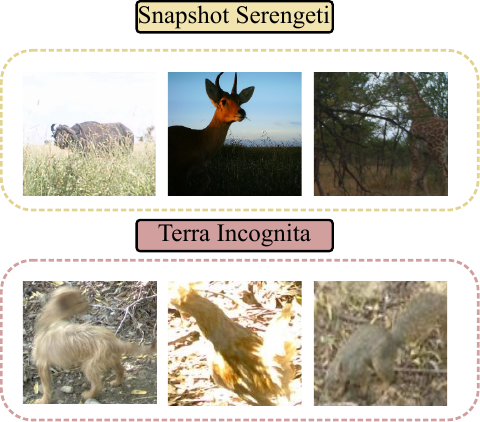}
    \caption{Cropped images from the Snapshot Serengeti and Terra Incognita datasets where we observe the domain shift and the difference in classes (different animal species).}
    \label{fig:datasets}
\end{figure}

\noindent \textbf{Datasets.}
We evaluate \method~using two public datasets in camera traps: Snapshot Serengeti \cite{dryad_5pt92} and Terra Incognita \cite{beery2018recognition}.
Some cropped images from these datasets are shown in \cref{fig:datasets}.
\begin{itemize}[leftmargin=*]
    \item \textit{Snapshot Serengeti} \cite{dryad_5pt92}. 
    We use the version of the Serengeti dataset used in WildCLIP \cite{gabeff2024wildclip}, which comprises $46$ classes.
    This dataset version includes $380 \times 380$ pixel image crops generated by the MegaDetector model from the Snapshot Serengeti project, applying a confidence threshold above $0.7$.
    Only camera trap images containing single animals were selected for this version.
    The dataset includes $340,972$ images divided into $230,971$ for training, $24,059$ for validation, and $85,942$ for testing.
    \item \textit{Terra Incognita} \cite{beery2018recognition}.
    This dataset comprises $16$ classes and introduces two testing groups called Cis-locations and Trans-locations.
    This means that the images taken from these locations are similar (Cis-locations) or different (Trans-locations) to the training data.
    These partitions were originally provided to test the robustness of computer vision models trained and evaluated in the same Terra Incognita dataset (in-domain evaluation).
    We filter the images in the dataset with the MegaDetector model of the library PyTorch-Wildlife \cite{hernandez2024pytorchwildlife}. 
    The dataset contains $45,912$ images divided into $12,313$ for training, $1,932$ for Cis-Validation, $1,501$ for Trans-Validation, $13,052$ for Cis-Test, and $17,114$ for Trans-Test.
\end{itemize}

\noindent \textbf{Evaluation protocol.}
We conduct two experiments to evaluate the performance of our model compared to the current state-of-the-art models.
In the first experiment, we use the Snapshot Serengeti dataset for training and validation, and the Terra Incognita dataset for testing (out-of-domain evaluation).
In other words, $\mathcal{D}$ and $\mathcal{S}$ are the Snapshot Serengeti and Terra Incognita datasets, respectively.
The Snapshot Serengeti dataset was collected in various protected areas in Africa, while Terra Incognita was collected in the American Southwest.
Therefore, we have two main problems in this experimental setup: i) the difference in data distribution between the two datasets $\mathcal{D}$ and $\mathcal{S}$ (domain shift), and ii) the difference in the sets of classes ($\mathcal{C}^D \neq \mathcal{C}^S$).
These two challenges are illustrated in \cref{fig:datasets}.
The difference in the sets of classes rules out any closed-set state-of-the-art method for comparison.
We report the accuracy results in the Cis-Test and Trans-Test sets of Terra Incognita.

For the second experiment, we modify our problem definition in Sec.~\ref{sec:Method} and use the same dataset to assess the model's performance without the complications introduced by the domain shift and new classes (in-domain evaluation).
More precisely, we use the Snapshot Serengeti or Terra Incognita datasets for training, validation, and testing.
This approach allows us to evaluate the model's accuracy and robustness within a consistent domain.

%To ensure optimal performance, hyperparameters were carefully selected using a grid search approach. This systematic exploration of parameter values allowed us to identify the best combination for each dataset.

%Hyperparameters for our model were carefully selected through a grid search approach to ensure optimal performance. This method systematically explores a range of parameter values, allowing us to identify the best combination for the given datasets.}
\vspace{0.1cm}

\noindent \textbf{Implementation details.} 
We use the 3.5 version of ChatGPT for the LLM, the ViT-B/16 version of CLIP, the 1.5-7B version of LLaVA, and the BERT-base-uncased version of BERT in our implementation of \method.
For training our model in the first experiment, we set $\alpha=0.6$ and $\tau=0.1$.
The MLP architecture consists of a single hidden layer with dimension $1,045$, and the Gaussian Error Linear Unit (GELU) as the activation function.
We train our model for $8$ epochs with a dropout rate of $0.27$.
We use the Stochastic Gradient Descent (SGD) algorithm to train \method~with a learning rate of $0.08$, momentum of $0.8$, and batch size of $48$.
For the second experiment, we fine-tuned \method~by unfreezing the CLIP image encoder and adjusting key hyperparameters: batch size of $100$, and training for $86$ epochs using SGD with a momentum of $0.8$. 
For the Snapshot Serengeti dataset, we use a $0.4$ dropout rate, $1\mathrm{e}{-3}$ learning rate, and an MLP with four hidden layers of $1,743$ dimensions.
For the Terra Incognita dataset, we use a $0.5$ dropout rate, $1\mathrm{e}{-4}$ learning rate, and an MLP with a single hidden layer of $1,045$ dimensions.
Early stopping was applied with a patience of $20$ epochs.
We optimize the hyperparameters of our model using random search.
%The code is publicly available at \url{https://github.com/Julian075/CATALOG}.

\subsection{Quantitative Results}

\noindent \textbf{Comparison with the state-of-the-art in out-of-domain evaluation.}
\cref{tab:zero-shot-results} reports the performance metrics of various models trained in datasets like TREEOFLIFE-10M and Snapshot Serengeti, and tested on the Cis-Test and Trans-Test sets of the Terra Incognita dataset, reflecting the models' accuracy (\%) in zero-shot learning scenarios.
We observe that CLIP ViT-B/32 achieves an accuracy of $32.18\%$ in Cis-Test and a Trans-Test accuracy of $26.62\%$, while CLIP ViT-B/16 improves these metrics to $39.14\%$ and $34.67\%$, respectively.
The WildCLIP model improves upon CLIP (ViT-B/16) with $40.38\%$ on Cis-Test and $38.90\%$ on Trans-Test due to its refinement in the Snapshot Serengeti dataset.
We also test the version Learning without Forgetting (LwF) of WildCLIP, which slightly alters performance with $41.60\%$ and $36.20\%$ on Cis-Test and Trans-Test, respectively.
In contrast, BioCLIP shows lower accuracy levels across both datasets, achieving $21.12\%$ on Cis-Test and $14.53\%$ on Trans-Test. 
\method~outperforms all previous FMs for camera-trap images, achieving $48.59\%$ of accuracy in Cis-Test and $41.92\%$ in Trans-Test.
These results highlight the advancements in zero-shot, domain-invariant, and open-set capabilities achieved by \method, surpassing previous state-of-the-art models.

\vspace{0.1cm}

\noindent \textbf{In-domain performance comparison in the Snapshot Serengeti dataset.}
\cref{tab:performance serengeti} shows a comparison of multiple models trained in the Snapshot Serengeti dataset.
CLIP-MLP is a modified version of CLIP's image encoder where we add an MLP and we train it with the regular cross-entropy loss.
This model achieves a test accuracy of $84.92\%$. 
However, it is worth noting that unlike WildCLIP and \method, the CLIP-MLP model lacks open vocabulary capabilities due to the cross-entropy training.
Therefore, CLIP-MLP is a strong baseline when no open-set capabilities are required.
WildCLIP performs moderately well with a test accuracy of $61.78\%$, but it lags behind \method.
The variant of WildCLIP, WildCLIP-LwF, shows a slight improvement, achieving a test accuracy of $64.39\%$.
The Learning without Forgetting (LwF) approach incorporated in this model appears to contribute positively, although the gain is not enough when compared to CLIP-MLP and \method.
\method~outperforms both CLIP-MLP and WildCLIP with $90.63\%$, showing its effectiveness for the case of in-domain evaluation.

\vspace{0.1cm}

\begin{table}[t]
\centering
\begin{tabular}{lcc}
\toprule
\textbf{Model} & \textbf{Loss Function} & \textbf{Test Acc (\%)} \\ \midrule
CLIP-MLP \cite{radford2021learning} & Cross-entropy & $84.92$ \\
\hdashline
WildCLIP \cite{gabeff2024wildclip} & Contrastive & $61.78$ \\ 
WildCLIP-LwF \cite{gabeff2024wildclip} & Contrastive & $64.39$ \\ 
\hdashline
%\method        & 66.14 \\ 
\textbf{\method~(ours)} & Contrastive  & $\textbf{90.63}$ \\ \bottomrule 
\end{tabular}
\caption{Performance comparison in Snapshot Serengeti (in-domain evaluation). All models use the ViT-B/16 backbone.}
\label{tab:performance serengeti}
\end{table}

\begin{table}[t]
\centering
\resizebox{\columnwidth}{!}{
\begin{tabular}{lcc}
\toprule
\textbf{Model} & \textbf{Cis-Test Acc(\%)} & \textbf{Trans-Test Acc(\%)} \\ \midrule
CLIP-MLP \cite{radford2021learning}  & $77.62$ & $ 71.88$ \\
\hdashline
%\method        & 61,364 & 53,472 \\ 
\textbf{WildCLIP} \cite{gabeff2024wildclip}  & $\textbf{91.72}$ & $\textbf{84.52}$ \\
WildCLIP-LwF \cite{gabeff2024wildclip}  & $88.96$ & $82.86$ \\
\hdashline
\method~(ours)  & $89.64$ & $84.32$ \\ \bottomrule
\end{tabular}
}
\caption{Performance comparison in the Terra Incognita dataset (in-domain evaluation). All models have the ViT-B/16 backbone.}
\label{tab:performance Terra}
\end{table}

\begin{table*}[h]
\centering
\def\arraystretch{0.8}
\setlength{\tabcolsep}{11pt}{
\begin{tabular}{cccccc}
\toprule
\textbf{CLIP} & \textbf{VLM} & \textbf{LLM} & \textbf{Templates} & \textbf{Cis-Test Acc (\%)} & \textbf{Trans-Test Acc (\%)} \\ \midrule
\xmark & \cmark & \xmark & \xmark & $2.54$ & $3.00$  \\
\xmark & \cmark & \xmark & \cmark & $8,37$  & $12,36$  \\
\xmark & \cmark & \cmark & \xmark & $11,20$ & $10,80$  \\
\xmark & \cmark & \cmark & \cmark & $14.51$ & $15.83$  \\ \hdashline
\cmark & \xmark & \xmark & \xmark & $39.14$  & $34.67$  \\ 
\cmark & \xmark & \cmark & \xmark & $37.92$  & $36.32$ \\
\cmark & \xmark & \xmark & \cmark & $44.26$ & $33.74$ \\
\cmark & \xmark & \cmark & \cmark & $44.39$ & $34.73$ \\ \hdashline
\cmark & \cmark & \cmark & \cmark & $\textbf{48.59}$ & $\textbf{41.92}$ \\
\bottomrule
\end{tabular}
}
\caption{Ablation studies for performance variations for different design choices of \method.
CLIP refers to the usage of the image encoder of CLIP. VLM refers to the usage of the Image-text Embedding module.
LLM refers to the description generated by ChatGPT for each animal species.
Finally, Templates refer to a set of predefined templates customized for our specific task in camera-trap image recognition. All models are trained on Snapshot Serengeti and evaluated on Terra Incognita (out-of-domain evaluation).}
\label{tab:ablation_study1}
\end{table*}

\noindent \textbf{In-domain performance comparison in the Terra Incognita dataset.}
\cref{tab:performance Terra} shows the performance of CLIP-MLP, WildCLIP, WildCLIP-LwF, and \method~when trained and evaluated on the Terra Incognita dataset.
The CLIP-MLP model performs well with a Cis-Test accuracy of $77.62\%$ and a Trans-Test accuracy of $71.88\%$.
Although this model performs well, its lack of open vocabulary capabilities remains a limitation.
WildCLIP achieves the highest Cis-Test accuracy of $91.72\%$ and the best Trans-Test accuracy of $84.52\%$. WildCLIP-LwF slightly underperforms WildCLIP, with a Cis-Test accuracy of $88.96\%$ and a Trans-Test accuracy of $82.86\%$. \method~achieves a Cis-Test accuracy of $89.64\%$, outperforming CLIP-MLP and WildCLIP-LwF, also performs competitively in the Trans-Test scenario with an accuracy of $84.32\%$, slightly lower than WildCLIP. Nevertheless, when we analyze the performance gap between the Cis-Test and Trans-Test in Terra Incognita, it is smaller in \method~ compared to other methods. This demonstrates that \method~ is effective at handling domain shifts, even within the same dataset.

\subsection{Ablation Studies}
\label{sec:ablations}

We conduct several ablation studies about i) the impact of the CLIP's image encoder, ii) the use of the VLM and the different textual descriptions of the categories, iii) the loss function used to train \method, and iv) the impact of using a text encoder capable of processing prompts longer than 77 tokens instead of BERT.

\vspace{0.1cm}

\noindent \textbf{CLIP image encoder, VLM, LLM, and templates.}
We evaluate the performance of \method~when we remove the VLM, CLIP image encoder, the descriptions generated by the LLM, and the set of predefined templates customized for the specific task in camera-trap image recognition.
\cref{tab:ablation_study1} shows the results in the Cis-Test and Trans-Test for different design choices in \method.
Our findings show that the lowest performance occurs when we remove CLIP, the LLM descriptions, and templates (first row in \cref{tab:ablation_study1}).
This is equivalent to setting $\alpha=0$ in \eqref{eqn:fusion} and using for text descriptions the base prompt ``A photo of a \{ \}". 
This results in scores of $2.54\%$ and $3.00\%$ in Cis-Test and Trans-Test, respectively. 
Similarly, removing CLIP and the templates or CLIP and the LLM descriptions also leads to very poor performance (second and third rows in \cref{tab:ablation_study1}).
When we remove CLIP and include VLM, the LLM descriptions, and the templates, we observe a little increase in performance, obtaining accuracies of $14.51\%$ in the Cis-Test and $15.83\%$ in the Trans-Test sets (fourth row in \cref{tab:ablation_study1}). These poor results highlight the important role of CLIP's image encoder in \method, meaning that our VLM alone cannot replace CLIP image encoder.

We observe a significant performance increase when we include the CLIP image encoder (fith row in \cref{tab:ablation_study1}), with scores of $39.14\%$ and $34.67\%$ in Cis-Test and Trans-Test, respectively.
This case reduces to the original CLIP model with the base prompt ``A photo of a \{ \}". 
When we incorporate the descriptions generated by the LLM (sixth row in \cref{tab:ablation_study1}), we obtain an increase in performance for the Trans-Test score ($36.32\%$) but a decrease for the Cis-Test score ($37.92\%$).
When we include the templates (seventh row in \cref{tab:ablation_study1}), we observe an increase in performance in Cis-Test ($44.26\%$) but a slight decrease in Trans-Test ($33.74\%$).
Finally, the combination of the LLM and template descriptions (eight row in \cref{tab:ablation_study1}) offers more robust results across Cis-Test ($44.39\%$) and Trans-Test ($34.73\%$).
These results suggest that including textual descriptions can be beneficial, but their effectiveness may vary depending on the test dataset and the choice of text information.

\cref{tab:ablation_study1} shows that the best performance is achieved by incorporating the VLM model, CLIP image encoder, the LLM descriptions, and templates (last row in \cref{tab:ablation_study1}), resulting in scores of $48.59\%$ (Cis-Test) and $41.92\%$ (Trans-Test).
This highlights the importance of integrating textual embedding techniques to better capture the relationships between images and their categorical descriptions.  The results also demonstrate how integrating FMs enhances the model's generalization performance. 
Furthermore, incorporating the image-text embeddings with the VLM and CLIP model provides an additional boost in accuracy, underscoring the effectiveness of this module for camera-trap image recognition.

\vspace{0.1cm}

\begin{table}[t]
\centering
\def\arraystretch{0.8}
\resizebox{\columnwidth}{!}{
\begin{tabular}{ccc}
\toprule
\textbf{Loss function} & \textbf{Cis-Test Acc(\%)} & \textbf{Trans-Test Acc (\%)} \\ \midrule
%\makecell{Supervised \\ Contrastive Loss} & 45.64 & 37.0193 \\
Sup. contrastive loss & $45.64$ & $37.02$ \\
\textbf{Contrastive loss} & $\textbf{48.59}$ & $\textbf{41.92}$ \\ \bottomrule
\end{tabular}
}
\caption{Ablation study on the choice of the loss function to train \method~in out-of-domain evaluation.}
\label{tab:ablation_study_loss}
\end{table}

\vspace{0.2cm}

\noindent \textbf{Evaluating different loss functions.} 
\cref{tab:ablation_study_loss} shows the comparison in performance of \method~when trained with the contrastive loss in \eqref{eqn:loss} and the well-known supervised contrastive loss defined in \cite{khosla2020supervised}.
The main difference between these two approaches is the elements we use as negative pairs.
We observe that \method~obtains $45.64\%$ in Cis-Test and $37.02\%$ in Trans-Test when trained with the supervised contrastive loss.
In contrast, our model achieves $48.59\%$ and $41.92\%$ in Cis-Test and Trans-Test, respectively when trained with the contrastive loss.
This indicates that the contrastive loss is effective in improving the model's ability to distinguish between different categories, leading to higher performance in both test scenarios.
Even though the supervised contrastive loss provides reasonable performance, it does not match the effectiveness of the standard contrastive loss for camera-trap image recognition.

% \begin{table}[t]
% \centering
% %\def\arraystretch{1.15}
% \resizebox{0.99\columnwidth}{!}{
% \begin{tabular}{ccc}
% \toprule
% \textbf{Text encoder} & \textbf{Cis-Test Acc(\%)} & \textbf{Trans-Test Acc (\%)} \\ \midrule
% %\makecell{Supervised \\ Contrastive Loss} & 45.64 & 37.0193 \\
% Long CLIP text encoder \cite{zhang2024long} & $28.55$ & $18.05$ \\
% CLIP \cite{radford2021learning} and BERT \cite{devlin2019bert} text encoder  & $\textbf{48.59}$ & $\textbf{41.92}$ \\ \bottomrule
% \end{tabular}
% }
% \caption{Ablation study on the choice of the text encoder to use \method~in out-of-domain evaluation.}
% \label{tab:ablation_study_text_encoder}
% \end{table}

\vspace{0.1cm}

\noindent \textbf{Evaluating text encoder.}
We evaluate the impact of using different text encoders on the performance of \method~in out-of-domain evaluation scenarios. 
Specifically, we compare the performance of the CLIP \cite{radford2021learning} and BERT \cite{devlin2019bert} text encoder combination against the Long CLIP text encoder \cite{zhang2024long}, which is capable of processing longer textual prompts (248 tokens). 
The study also uses the Long CLIP image encoder, ensuring a consistent comparison between the models. 
All hyperparameters were kept unchanged and optimized based on the values found for the out-of-domain evaluation.

\cref{tab:ablation_study_text_encoder} shows the Cis-Test and Trans-Test accuracy results for the two encoder setups. The CLIP and BERT text encoder combination achieved the highest performance, with Cis-Test and Trans-Test accuracies of 48.59\% and 41.92\%, respectively. In contrast, using the Long CLIP text encoder significantly reduced performance, achieving only 28.55\% accuracy on the Cis-Test and 18.05\% on the Trans-Test.
This performance drop suggests that simply increasing the text encoder's capacity to process more tokens does not guarantee improved alignment and performance in out-of-domain scenarios.

\begin{table}
\centering
\resizebox{\columnwidth}{!}{
\begin{tabular}{cccc}
\toprule
\textbf{Long CLIP} & \textbf{CLIP$+$BERT} & \textbf{Cis-Test Acc(\%)} & \textbf{Trans-Test Acc (\%)} \\ \midrule
%\textbf{Text encoder} & \textbf{Cis-Test Acc(\%)} & \textbf{Trans-Test Acc (\%)} \\ \midrule
%\makecell{Supervised \\ Contrastive Loss} & 45.64 & 37.0193 \\
\cmark & \xmark & $28.55$ & $18.05$ \\
\xmark & \cmark & $\textbf{48.59}$ & $\textbf{41.92}$ \\ \bottomrule
\end{tabular}
}
\caption{Ablation study on the choice of the text encoder to use in \method~for out-of-domain evaluation.}
\label{tab:ablation_study_text_encoder}
\end{table}

\subsection{Sensitivity to the Hyperparameter $\alpha$}

\begin{figure}
    \centering
    \begin{subfigure}[t]{0.49\columnwidth}
        \centering
        \includegraphics[width=\textwidth]{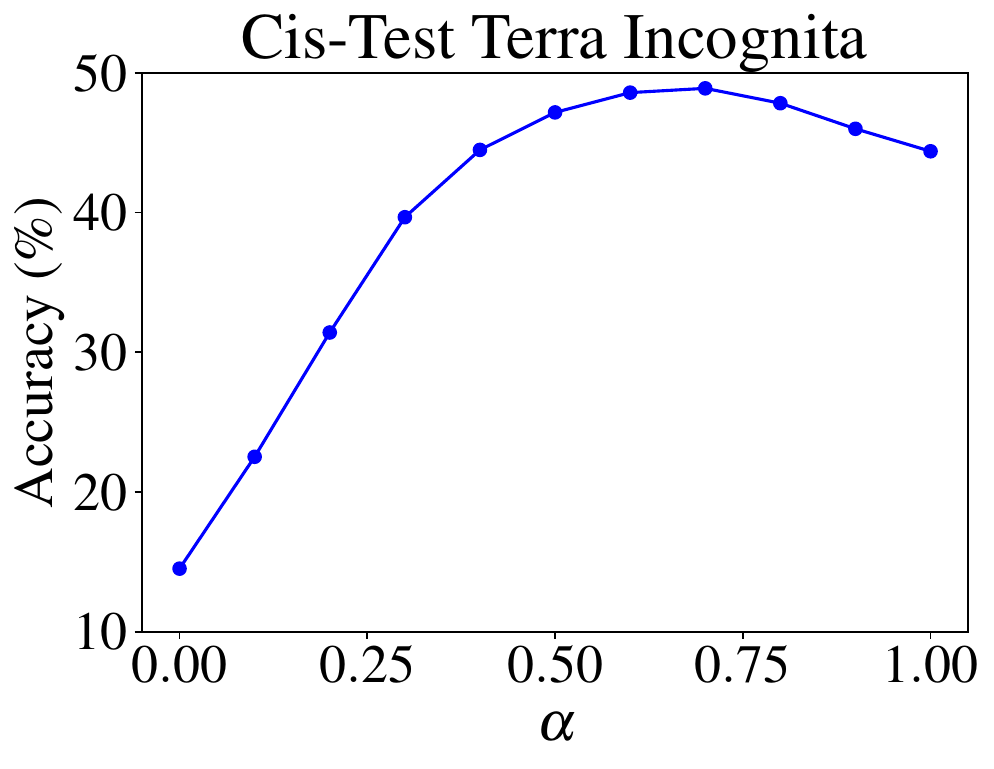}
        \caption{Cis-Test set.}
        \label{fig:alpha cis}
    \end{subfigure}%
    \hfill
    \begin{subfigure}[t]{0.49\columnwidth}
        \centering
        \includegraphics[width=\textwidth]{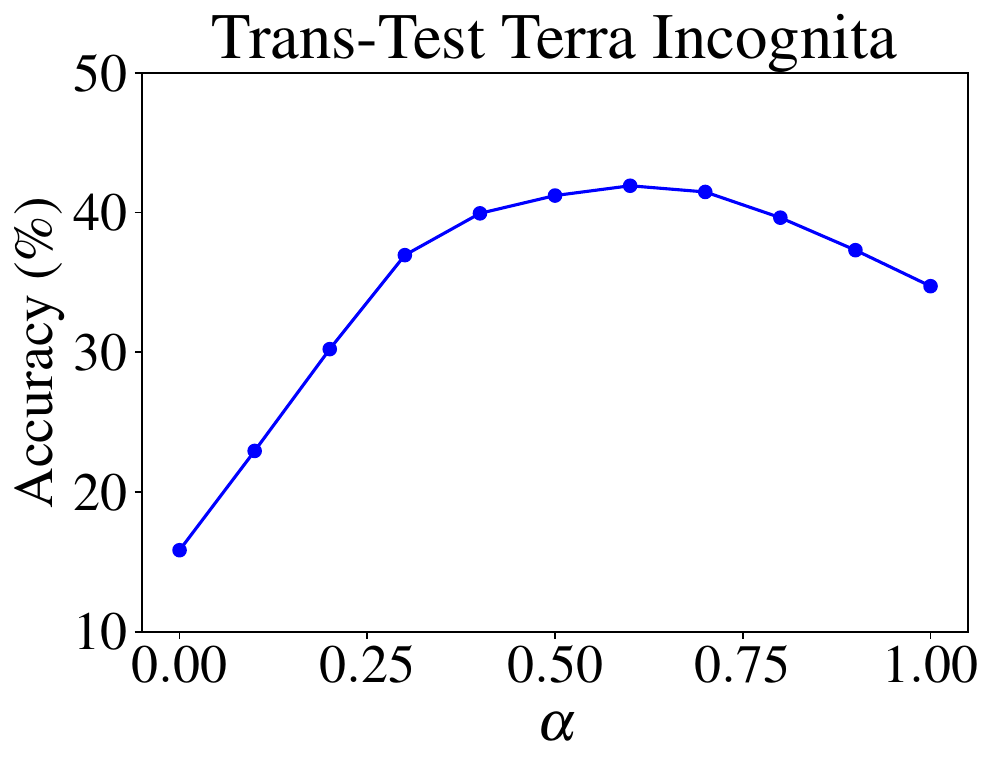}
        \caption{Trans-Test set.}
        \label{fig:alpha trans}
    \end{subfigure}
    \caption{Sensibility analysis of the hyperparameter $\alpha$ of \method~in the Terra Incognita dataset for out-of-domain evaluation.}
    \label{fig:sensibility}
\end{figure}

Fig.~\ref{fig:alpha cis} and \ref{fig:alpha trans} show the change of \method's performance to variations of the parameter $\alpha$ between $0$ and $1$ in \eqref{eqn:fusion} for Cis-Test and Trans-Test in Terra Incognita for out-of-domain evaluation.
We observe that the information from both matrices $\mathbf{Q}$ and $\mathbf{W}$, are complementary.
Both Fig.~\ref{fig:alpha cis} and \ref{fig:alpha trans} show that the optimal value for $\alpha$ is $0.6$.
This indicates that giving nearly equal importance to both matrices provides the best accuracy.
When $\alpha$ deviates from the optimal value, the accuracy decreases, suggesting that overemphasizing either matrix leads to a loss of valuable information for classification.
This consistency across both evaluation sets highlights the robustness of the model's performance when information from both matrices is used.

\subsection{Limitations}

Although FMs have shown promising results in recognizing animal species in camera-trap images, there is still a noticeable difference between their performance on out-of-domain (\cref{tab:zero-shot-results}) and in-domain (\cref{tab:performance Terra}) data evaluation in Terra Incognita.
This highlights the need to enhance the generalization capabilities of these models for camera-trap image recognition.
%This suggests that there is room for improvement in the generalization capabilities of these models for camera-trap image recognition.
In the supplementary material, we provide a more detailed analysis of the limitations of the CATALOG model.
Two potential directions for addressing this issue include: i) collecting a large-scale dataset of image-text pairs of camera-trap images to train a foundation model like CLIP, and ii) leveraging expert knowledge through textual information about each animal species to fill the domain gap.
These ideas warrant further exploration and are left for future work.

\section{Conclusions}
\label{sec:conclusions}
In this paper, we introduced \method, a new model that integrates multiple FMs to address the performance loss caused by domain changes in camera-trap image recognition. 
\method~tackles domain shifts by using robust features extracted by CLIP, LLaVA, and BERT, combined with stronger category descriptions generated by an LLM and predefined templates specific to the camera-trap context.
Our extensive experiments show that \method~outperforms state-of-the-art models in camera-trap image classification, especially in the case when there are domain shifts between the training and testing dataset, all while maintaining its open vocabulary capabilities.

\vspace{0.1cm}

\noindent \textbf{Acknowledgments.}
This work was supported by Universidad de Antioquia - CODI and Alexander von Humboldt Institute for Research on Biological Resources (project 2020-33250), and by the ANR (French National Research Agency) under the JCJC project DeSNAP (ANR-24-CE23-1895-01).
%%%%%%%%% REFERENCES
{\small
\bibliographystyle{ieee_fullname}
\bibliography{egbib}
}

\end{document}

%\title{Language-guided CLIP Adaptation for Zero-shot Camera Trap Image Recognition}
%\title{Bridging the Gap between Foundation Models and Camera Trap Image Recognition: supplementary}
\title{\method: A Camera Trap Language-guided Contrastive Learning Model}
\author{
Julian D. Santamaria$^1$, Claudia Isaza$^1$, Jhony H. Giraldo$^2$ \\
$^1$ SISTEMIC, Faculty of Engineering, Universidad de Antioquia-UdeA, Medellín, Colombia. \\
$^2$ LTCI, Télécom Paris, Institut Polytechnique de Paris, Palaiseau, France. \\
{\tt \small \{julian.santamaria, victoria.isaza\}@udea.edu.co, jhony.giraldo@telecom-paris.fr}
}
\maketitle

\appendix 
This supplementary material includes some examples of the predefined templates and detailed examples of specific prompts used to generate the category descriptions with the LLM \cite{brown2020language}. 
Additionally, it provides the prompt used to create textual descriptions of animals present in cropped images using LLaVA \cite{liu2024visual}.
%This supplementary includes detailed examples of the specific prompts used to generate category descriptions with ChatGPT and predefined templates. Additionally it provides the prompts employed to create textual descriptions of the animals present in the cropped images using LLaVA.

\section{Templates}
In this section, we present some examples of templates designed for the camera-trap image recognition task. These templates have been adapted from the ImageNet templates used in CLIP \cite{radford2021learning} and are shown below:
%In this section, we provided the templates tailored to our specific task of camera-trap image recognition. The templates are based in the ImageNet templates used in CLIP \cite{radford2021learning} and are the follows: \\
\begin{itemize}
    \small
    \itemsep0em 
    \item a photo captured by a camera trap of a \{\}.
    \item a camera trap photo of the \{\} captured in poor conditions.
    \item a cropped camera trap image of the \{\}.
    \item a camera trap image featuring a bright view of the \{\}.
    \item a camera trap image of the \{\} captured in clean conditions.
    \item a camera trap image of the \{\} captured in dirty conditions.
    \item a camera trap image with low light conditions featuring the \{\}.
    \item a black and white camera trap image of the \{\}.
    \item a cropped camera trap image of a \{\}.
    \item a blurry camera trap image of the \{\}.
    \item a camera trap image of the \{\}.
    \item a camera trap image of a single \{\}.
    \item a camera trap image of a \{\}.
    \item a camera trap image of a large \{\}.
    \item a blurry camera trap image of a \{\}.
    \item a pixelated camera trap image of a \{\}.
    \item a camera trap image of the weird \{\}.
    \item a camera trap image of the large \{\}.
    \item a dark camera trap image of a \{\}.
    \item a camera trap image of a small \{\}.
\end{itemize}
For each template, we replace ``\{ \}" by the specific category in $\mathcal{C}^D$.

\section{Prompts}

In this section, we provide the prompts for the LLM and LLaVA used in \method.
%used to generate the different descriptions used in \method~ pipeline.
%In this section, we outline the prompts utilized to generate various descriptions within the \method~ pipeline.

\subsection{Prompt LLM}
The prompt utilized to get the LLM description of the animal species follows a structure based on the methodology discussed in \cite{pratt2023does} and is shown below.\\
\noindent\fbox{%
    \parbox{\columnwidth}{%
        You are an AI assistant specialized in biology and providing accurate and detailed descriptions of animal species. We are creating detailed and specific prompts to describe various species. The goal is to generate multiple sentences that capture different aspects of each species' appearance and behavior. Please follow the structure and style shown in the examples below. Each species should have a set of descriptions that highlight key characteristics.\\\\
        Example Structure:\\\\
        Badger: 
        \begin{itemize}
            \small
            \itemsep0em 
            \item a badger is a mammal with a stout body and short sturdy legs.
            \item a badger's fur is coarse and typically grayish-black.
            \item badgers often feature a white stripe running from the nose to the back of the head dividing into two stripes along the sides of the body to the base of the tail.
            \item badgers have broad flat heads with small eyes and ears.
            \item badger noses are elongated and tapered ending in a black muzzle.
            \item badgers possess strong well-developed claws adapted for digging burrows.
            \item overall badgers have a rugged and muscular appearance suited for their burrowing lifestyle.
        \end{itemize}
    }%
}
\begin{figure*}
    \centering
    \includegraphics[width=1.5\columnwidth]{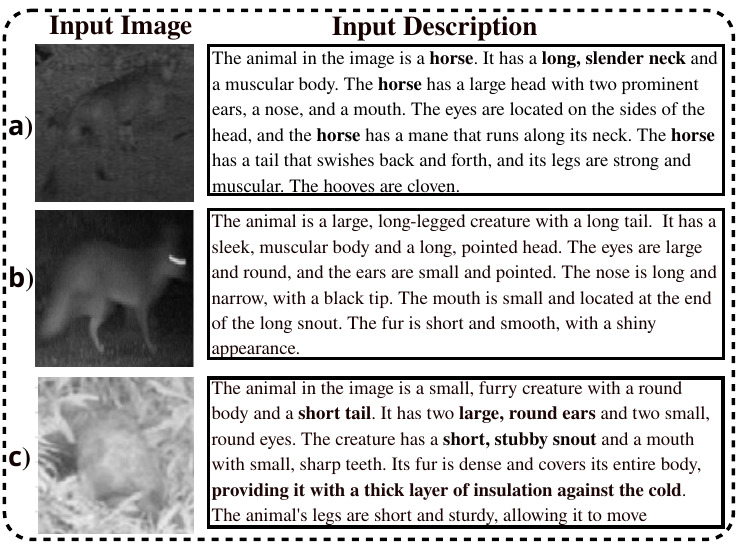}
    \caption{Failure cases of the CATALOG model for camera trap image classification. \textbf{case a:} The VLM generates an incorrect description with details that do not match the input image. \textbf{case b:} A blurry image results in a vague and unhelpful description. \textbf{case c:} When the input image is very unclear, the VLM generates a random and irrelevant description. These examples show that when the descriptions are wrong or not informative, the model makes incorrect predictions.}

    %\caption{The performance of the foundation models for wildlife monitoring BioCLIP \cite{stevens2024bioclip} and WildCLIP \cite{gabeff2024wildclip} drops when evaluated on the Terra Incognita dataset \cite{beery2018recognition}.}
    \label{fig:failure_cases}
\end{figure*}
%The prompt employed is the show follow and the structure are based of \cite{pratt2023does}.\\

\subsection{Prompt LLaVA}
The prompt used in LLaVA aligns with the approach employed in \cite{fabian2023multimodal} and is structured as follows:\\
%The prompt employed in LLaVA aligns with the one utilized in \cite{fabian2023multimodal} and is as follows:\\
\noindent\fbox{%
    \parbox{\columnwidth}{%
        \small
        \textbf{SYSTEM}: You are an AI assistant specialized in biology and providing accurate and detailed descriptions of animal species.\textbackslash n $\ll 
        \text{image} \gg$ \textbackslash n\\
        \textbf{USER}: You are given the description of an animal species. Provide a very detailed description of the appearance of the species and describe each body part of the animal in detail. Only include details that can be directly visible in a photograph of the animal. Only include information related to the appearance of the animal and nothing else. Make sure to only include information that is present in the species description and is certainly true for the given species. Do not include any information related to the sound or smell of the animal. Do not include any numerical information related to measurements in the text in units: m cm in inches ft feet km/h kg lb lbs. Remove any special characters such as unicode tags from the text. Return the answer as a single paragraph.
    }%
}

% ``[SYSTEM] You are an AI assistant specialized in biology and providing accurate and detailed descriptions of animal species.\textbackslash n $\ll image \gg$ \textbackslash nUSER: You are given the description of an animal species. Provide a very detailed description of the appearance of the species and describe each body part of the animal in detail. Only include details that can be directly visible in a photograph of the animal. Only include information related to the appearance of the animal and nothing else. Make sure to only include information that is present in the species description and is certainly true for the given species. Do not include any information related to the sound or smell of the animal. Do not include any numerical information related to measurements in the text in units: m cm in inches ft feet km/h kg lb lbs. Remove any special characters such as unicode tags from the text. Return the answer as a single paragraph.\textbackslash nASSISTANT:"

\section{Analysis of \method's Errors}
In this section, we provide a more detailed analysis of the limitations of the CATALOG model. Fig.~\ref{fig:failure_cases} illustrates how sensitive the model is to the input descriptions generated for the VLM. These descriptions provide additional information to help the model make better classifications. However, when the descriptions are wrong or unclear, the model makes incorrect predictions.

To clarify the model's sensitivity to input descriptions, Fig.~\ref{fig:failure_cases} presents three examples:
\begin{itemize}
    \item In \textbf{case a}, the VLM hallucinates by adding details that are not in the image. For example, it describes the animal as a horse, even though the input image does not match this description. This leads to a completely incorrect prediction.

    \item In \textbf{case b}, the input image is blurry and hard to interpret. The VLM generates a vague description with little useful information. As a result, the model does not get enough context to make the correct prediction.

    \item In \textbf{case c}, the image is so unclear that the VLM creates a random description that has no connection to the input. This random description further confuses the model, leading to a wrong prediction.
\end{itemize}

These examples show that the CATALOG model depends on the accuracy and quality of the descriptions created by the VLM. When the descriptions are not reliable or informative, the model struggles to classify the input correctly. Improving the robustness of the VLM is crucial for handling noisy or unclear inputs.

%%%%%%%%% REFERENCES
{\small
\bibliographystyle{ieee_fullname}
\bibliography{egbib}
}